\pdfoutput=1

\documentclass[11pt]{article}

\usepackage{acl}
\usepackage{times}
\usepackage{latexsym}
\usepackage{paralist}

\usepackage[T1]{fontenc}

\usepackage[utf8]{inputenc}

\usepackage{microtype}
\usepackage{graphicx}
\usepackage[export]{adjustbox}

\usepackage{booktabs}

\usepackage[appendix=strip]{apxproof}

\title{Extracting Definienda in Mathematical Scholarly Articles with~Transformers}

\author{Shufan Jiang \\
  DI ENS, ENS, PSL University, CNRS \\
  \& Inria \\
  Paris, France \\
  \texttt{shufan.jiang@ens.psl.eu} \\\And
  Pierre Senellart \\
  DI ENS, ENS, PSL University, CNRS \\
  \& Inria \& Institut Universitaire de France\\
  Paris, France\\
  \texttt{pierre@senellart.com} \\}

\begin{document}
\maketitle
\begin{abstract}
We consider automatically identifying the defined term within a
mathematical definition from the text of an academic article. Inspired by
the development of transformer-based natural language processing
applications, we pose the problem as (a)~a token-level classification
task using fine-tuned pre-trained transformers; and~(b) a question-answering task using a generalist large language model (GPT). We also propose a rule-based approach to build a labeled dataset from the \LaTeX{} source of papers. Experimental results show that it is possible to reach high levels of precision and recall using either recent (and expensive) GPT~4 or simpler pre-trained models fine-tuned on our task.
\end{abstract}

\section{Introduction}
Mathematical scholarly articles contain mathematical statements such as axioms, theorems, proofs, etc. These structures are not captured by traditional ways of navigating the scientific literature, e.g., keyword search. We consider initiatives aiming at better knowledge discovery from scientific papers
such as s\TeX{}~\citep{kohlhase2010stex}, a bottom-up solution for mathematical knowledge management that relies on authors adding explicit metadata when writing in \LaTeX;
MathRepo~\citep{fevola2022mathrepo}, a crowd-sourced repository for mathematicians to share any additional research data alongside their papers;
or TheoremKB~\citep{mishra_towards_2021}, a project that extracts the location of theorems and proofs in mathematical research articles.
Following these ideas, we aim at automatically building a knowledge graph to automatically index articles with the terms defined therein.

\begin{figure}[t]
    \centering
    \includegraphics[width=\linewidth, frame]{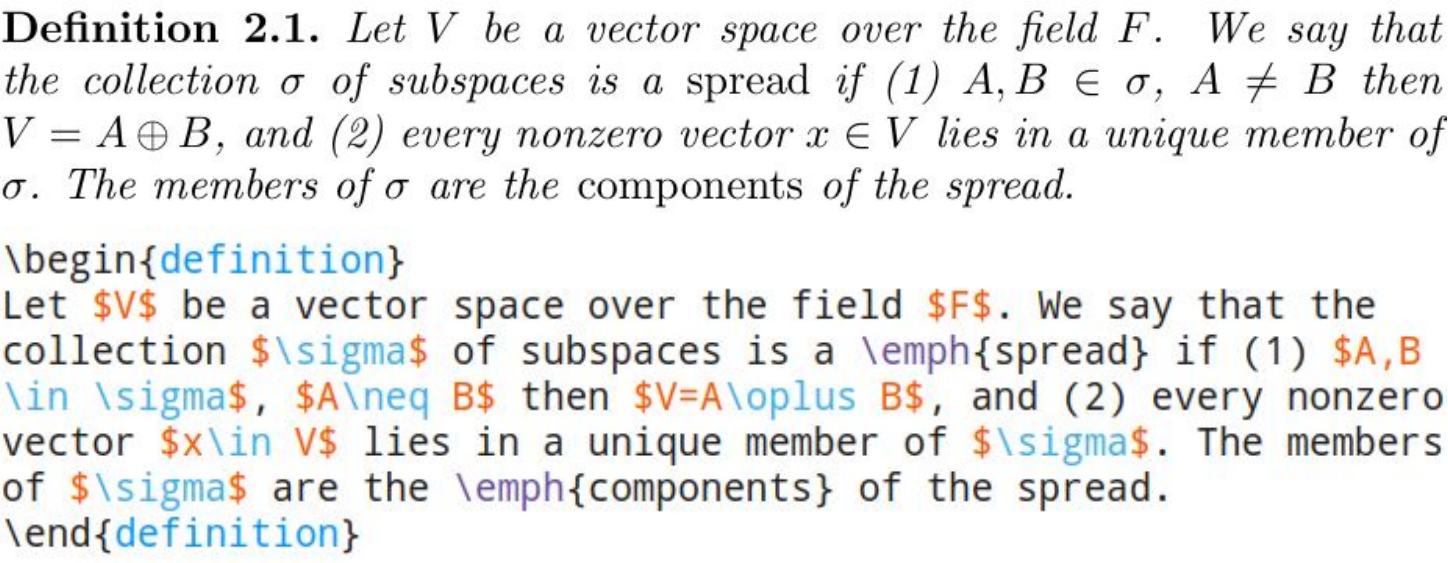}
    \caption{Rendering of a definition from a mathematical scholarly
    article~\citep{Nagy_2013} accompanied with its \LaTeX{} source code. The definienda are ``spread'' and ``components''.}
    \label{fig:def_spread}
\end{figure}

As a first step, we consider the simpler problem of, given the text of a
formal mathematical definition (which is typically obtained from the
PDF article), extracting the \emph{definienda} (terms defined within). As
an example, we show in Figure~\ref{fig:def_spread} a mathematical
definition (as rendered within a PDF article, accompanied with its
\LaTeX{} source code) that defines two terms (which we call the \emph{definienda}):
``spread'' and ``components''. In this particular example, the two terms
are \emph{emphasized} in the PDF (by being set in a non-italic font within an italic
paragraph) -- this is not always the case but we will exploit the fact
that some authors do this to build a labeled dataset of definitions and
definienda.

After discussing some related work in Section~\ref{sec:related}, we describe our approach in Section~\ref{sec:approach} and show experimental results in Section~\ref{sec:eval}.

\section{Related Work}
\label{sec:related}
The difficulties of our task lie in (1)~the lack of labeled datasets; (2)~the diversity in mathematicians' writing style; and (3)~the interplay of discourse and formulae, which differentiate mathematical text and text in the general domain. We review potential corpora and existing approaches in this section.

The most relevant work to our objective is by
\citet{berlioz2023hierarchical}.  The author trains supervised
classifiers to extract definitions from mathematical papers from arXiv.
The best classifier takes static word embeddings built from arXiv papers,
part-of-speech features of the words, and hand-coded binary features,
such as if a word is an acronym, and then applies a BiLSTM-CRF
architecture for sequence tagging~\citep{huang2015bidirectional}. The
resulting precision, recall, and F$_1$ are of 0.69, 0.65, and 0.67
respectively. The author uses the classifier to automatically extract
term-definition pairs from arXiv articles and Wikidata, resulting in the
dataset ArGot~\citep{berlioz_argot_2021}. Note however that a limitation
of ArGot, which makes it unsuitable in our setting, where the text of definitions is directly taken from PDFs, is that mathematical expressions and formulas are masked out in the training set.

Another related task is term-definition extraction in the general domain
of scientific articles. For example, Scholarphi
\citep{head2021augmenting} is an augmented reading interface for papers
with publicly available \LaTeX{} sources. Given a paper (with its
\LaTeX{} source), it lets the reader click on specific words to view
their definitions within the paper. The authors test several models for
definition--term detection, including an original Heuristically Enhanced
Deep Definition Extraction \citep{kang_document-level_2020},
syntactic features, heuristic rules, and different word representation
technologies such as contextualized word representations based on
transformers~\citep{vaswani2017attention}. The results show that models involving SciBERT~\citep{beltagy2019scibert} achieved higher accuracy on most measurements due to the domain similarity between the scholarly documents for pre-training SciBERT and those used in the evaluation.
 Following this idea, cc\_math\_roberta~\citep{mishra2023multimodal} is a
 RoBERTa-based model pertained from scratch on mathematical articles from
 arXiv~\citep{mishra2023multimodal}. This model outperforms Roberta in a sentence-level classification task while the corpora size for pre-training cc\_math\_roberta is much smaller than Roberta's. We aim to determine in this work if contextualized word representations can improve the results of mathematical definienda extraction.

NaturalProofs~\citep{welleck2021naturalproofs} is a corpus of mathematical
statements and their proofs. These statements are extracted from
different sources with hand-crafted rules, such as the content being enclosed by \verb|\begin{theorem}| and \verb|\end{theorem}| in the \LaTeX{} source of a textbook project on algebraic stacks\footnote{\url{https://github.com/stacks/stacks-project}}.  Each statement is either a theorem or a definition. However, this dataset does not annotate the definienda of each definition.

\section{Proposed Approach}
\label{sec:approach}
We describe our approach in two steps.
    First, we build a ground-truth dataset using the \LaTeX{} source of
    papers. As the existing large datasets either concern
    term--definition extraction from general corpora like web pages or
    textbooks~\citep{welleck2021naturalproofs} or mask out mathematical
    expressions in the text~\citep{berlioz_argot_2021}, we decide to
    process plain text as it appears in scholarly papers so that our
    solution can be directly applied to texts extracted from PDF articles when the \LaTeX{} source is unavailable.
    Second, we study different usages of transformer-based models to extract definienda. We are interested in fine-tuning and one-shot learning (prompt engineering).
The source code of our approach, as well as the constructed dataset,  is available on Github\footnote{\url{https://github.com/sufianj/def_extraction}}.
\subsection{Dataset Construction}

To start with a reasonable corpus, we collected the \LaTeX{} source of
all 28\,477 arXiv papers in the area of Combinatorics (arXiv's
\textsf{math.CO} category) published before 1st Jan 2020 through arXiv’s
bulk access from
AWS\footnote{\url{https://info.arxiv.org/help/bulk_data_s3.html}}. Our
goal in building the dataset was \emph{not} to be complete, but to
produce as cheaply and reliably as possible a ground-truth dataset of
definitions and definienda. For this purpose, we rely on two features of
definitions that some authors (but definitely not all!) use: definienda
are often written in italics within the definition (or, as in
Figure~\ref{fig:def_spread}, in non-italics within an italics paragraph);
and definienda are sometimes shown in parentheses after the definition
header. As we do not need to completely capture all cases in the building
of the dataset, we assume that definitions are within a
\texttt{definition} \LaTeX{} environment and thus extracted text blocks
between \verb|\begin{definition}| and  \verb|\end{definition}|;
  we ignored contents enclosed in other author-defined environments, such as \verb|\begin{Def}|, which might bring us more definitions but also more noise.
For defined terms, relying on the two features described above, we extracted the contents within \verb|\textit{}| and \verb|\emph{}| from the text blocks as well as the content potentially provided as optional argument to the  \verb|\begin{definition}[]| environment.
    We then converted the extracted partial \LaTeX{} code into plain text with Unicode characters using \texttt{pylatexenc}\footnote{\url{https://github.com/phfaist/pylatexenc}}. After a brief glance at the most frequent extracted definienda values, we handcrafted regular expressions to filter out the following recurrent noises among them:
\begin{compactitem}
    \item irrelevant or meaningless phrases such as repeating ``i.e.'' and ``\verb|\d|'';
    \item Latin locutions such as ``et al.'';
    \item list entries such as ``(i)'' and ``(iii)''.
\end{compactitem}
After filtering, we got a list of 13\,692 text blocks, of which the
average length is 70 tokens, and the maximum length is 5\,266 tokens. We removed 39 text blocks having more than 500 tokens.
Finally, we labeled automatically the texts with IOB2 tagging, where the ``B-MATH\_TERM'' tag denotes the first token of every defined term, ``I-MATH\_TERM'' tag indicates any non-initial token in a defined term, and the ``O'' tag means that the token is outside any definiendum. Considering partially italics compound terms like ``\verb|\emph{non}-k-equivalent|'', we annotate ``non-k-equivalent'' as a definiendum. We sorted the labeled texts by the last update time of the papers.

To evaluate the quality of this dataset, we examined by hand 1\,024
labeled entries. We found that only 30 annotated texts out of 1\,024 to
be incorrectly labeled, confirming the quality of our annotation.  We manually removed or corrected wrong annotations and got 999 labeled texts, which became our ground truth test data. We built training/validation sets for 10-fold cross-validation with the rest of the labeled texts, to separate them from our test data.

\subsection{Fine-tuning Pre-trained Language Models for Token Classification}
\label{sec:fine-tuning}
For the fine-tuning setup, we consider the extraction of definienda as a
token-level classification problem: given a text block, the classifier
labels each token as B-MATH\_TERM, I-MATH\_TERM or O. We used the
implementation for token classification
\textit{RobertaForTokenClassification} in the transformers
package~\citep{wolf-etal-2020-transformers}. It loads a pre-trained
language model and adds a linear layer on top of the token representation
output. We experimented with an out-of-the-box and general language model
Roberta-base~\citep{liu2019roberta} and a domain-specific model
cc\_math\_roberta~\citep{mishra2023multimodal}. Since
\citet{mishra2023multimodal} do not report
performance on token-level tasks, we
used two checkpoints of it, one pretained for 1 epoch (denoted as
cc\_ep01)\footnote{\url{https://huggingface.co/InriaValda/cc_math_roberta_ep01}},
and another pre-trained for 10 epochs (denoted as
cc\_ep10)\footnote{\url{https://huggingface.co/InriaValda/cc_math_roberta_ep10}}.
Then we fed the 10 train/validation sets to train the linear layer to
predict the probability of a token's representation matching one of the
three labels. We set the maximum sequence length of the model to 256. We
ran all our experiments with a fixed learning rate of $5\cdot 10^{-5}$ and a fixed
batch size of 16. We searched the best number of epochs among [3, 5, 10].
We also experimented with 1\,024, 2\,048, and 10\,240 samples from each training set to see the performance of the classifiers with low resources. As Roberta-base and cc\_math\_roberta have their own tokenizers, the models' output loss and accuracy are based on different numbers of word pieces and are not comparable. To evaluate the predictions, we used the predicted tag of the first word piece of each word and regrouped the IOB2-tagged word into definienda. We present our unified evaluation over ground truth data in Section~\ref{sec:eval}.

\subsection{Querying GPT}
Driven by the growing popularity of few-shot learning with pre-trained language models~\citep{brown_language_2020}, we also query the GPT language model, using different available versions: we first experimented with ChatGPT\footnote{An example of our conversation: \url{https://chat.openai.com/share/c96b156f-cba1-4804-8f19-1622a9bc564e}} (based on GPT 3.5) and then used the API versions of GPT-3.5-Turbo and GPT-4. We initially gave ChatGPT only one example in our question and attempted to obtain a IOB2-compliant output. We quickly realized that the returned tagging was random, unstable, and incoherent with the expected terms. However, if we ask ChatGPT to return the definienda directly, we get more pertinent results. We thus asked GPT-3.5-Turbo and GPT-4 to identify the definienda in our ground truth data via OpenAI's API. %
For each request, we send the same task description (system input) and  a
text from our test data (user input).  We fixed the max output length to
128 and temperature to 0. By the time of writing, the cost of these API
are count by tokens -- GPT-4 8K context model's input and output token
prices are 20 and 30 times that of GPT-3.5 4K context model. Since GPT-4 tend to give more precise and shorter responses, the cost of GPT-4 on our task is roughly 20 times that of GPT-3.5. For our test, we spent \$0.42 on GPT-3.5 and \$7.80 on GPT-4.

\section{Evaluation}
\label{sec:eval}
Now that we got the predictions from our fine-tuned token classifiers and
the answers from GPT models, we compared them with ground truth data. We
first removed the repeated expected definienda for each annotated text
and got 1\,552 unique definienda in total. Then we converted both
expected terms and extracted terms to lowercase. For each unique expected
term, if it is the same as an extracted term, we counted one ``True
Positive''. We counted one ``Cut Off'' if it contains an extracted term.
If it is contained in an extracted term, we counted one ``Too Long''.
Finally, we removed all spaces in the expected term to make an expected
no-space string, and we joined all extracted terms to make an extracted
no-space string; if the extracted no-space string contains the expected
no-space string, we considered that the expected term is extracted as one
``True Positive or Split Term''. We calculated the precision, recall, and
F$_1$-score using the ``True Positive or Split Term'' count to have a
higher tolerance for boundary errors on all models.
Table~\ref{tab:eval_gpt} shows the results of GPT's answers.
Tables~\ref{tab:eval_10_fold_2048} and~\ref{tab:eval_10_fold_10240}
present the averaged performance of cc\_ep01, cc\_ep10 and Roberta over
10-fold cross-validation. We set the best precision, recall, and
F$_1$-scores in bold across these three tables.

Our first remark is the high recall of GPTs' answers. Indeed, %
GPT models, especially GPT-3.5, tend to return everything in the given
text, resulting in poor precision. After checking the outputs over the
1024 test data, we found an over-prediction of formulas and mathematical
expressions, which corresponds to the analysis by~\citep {kang_document-level_2020}.

Our second remark is that fine-tuned classifiers have more balanced
precision and recall, as the numbers of extracted terms are closer to the
expected number (1\,552). To our surprise, although the tokenizer of cc\_math\_roberta models produced fewer word pieces than Roberta's tokenizer, Roberta-base yielded the best performance among the three models in our task, regardless the size of the training set. Moreover, cc\_math\_roberta models' performance varies more than Roberta's (see in Table~\ref{tab:eval_10_fold_std_f1}), showing that cc\_math\_roberta models are less robust to different input data.

In all the setups, cc\_ep01 was always the worst for our task, implying
the benefit of pre-training. The performances of all fine-tuned models
improve significantly as the training set size increases. When given 10\,240 training data, fine-tuning a pre-trained model gives better overall predictions than GPT-4, and when given 2048 training data, fine-tuned Roberta-base already gives better precision than GPT-4.

Finally, note that these finetuned language models are obviously much
less computationally expensive than OpenAI's GPT models.

\begin{table}[t]
    \centering
    \begin{tabular}{lrr}
    \toprule
        \textbf{Model} & \textbf{GPT-3.5} & \textbf{GPT-4} \\ \midrule
        Extracted & \textcolor{red}{6867} & 2245 \\
        True Positive & 1072 & 942 \\
        TP+Split Term & 1315 & 1383 \\
        Too Long & 379 & 595 \\
        Cut Off & 656 & 138 \\
        Precision & \textcolor{red}{0.1929} & 0.6248 \\
        Recall & \textbf{0.8312} & \textbf{0.8821} \\
        F$_1$ & 0.3131 & \textbf{0.7315} \\ \bottomrule
    \end{tabular}
    \caption{\label{tab:eval_gpt}Performance comparison of extraction by GPT models. The huge number of extracted terms results in the poor precision of GPT-3.5 model.}
\end{table}

\begin{table}[t]
    \centering
    \begin{tabular}{lrrr}
    \toprule
        \textbf{Model} & \textbf{cc\_ep01} & \textbf{cc\_ep10} & \textbf{Rob.} \\ \midrule
        Extracted & 2093.0 & 1710.8 & 1764.2 \\
        True positive & 514.9 & 881.2 & 934.2 \\
        TP+Split Term & 693.8 & 1056.5 & 1127.5 \\
        Too Long & 170.2 & 209.1 & 268.8 \\
        Cut Off & 522.6 & 405.2 & 326.1 \\
        Precision & 0.354 & 0.623 & 0.646 \\
        Recall & 0.447 & 0.681 & 0.726 \\
        F$_1$ & 0.383 & 0.647 & 0.679 \\ \bottomrule
    \end{tabular}
    \caption{\label{tab:eval_10_fold_2048}Averaged performance of fine-tuned models, with 2048 training data.}
\end{table}

\begin{table}[t]
    \centering
    \begin{tabular}{lrrr}
    \toprule
        \textbf{Model} & \textbf{cc\_ep01} & \textbf{cc\_ep10} & \textbf{Rob.} \\ \midrule
        Extracted & 1775.2 & 1779.2 & 1770.5 \\
        True positive & 540.3 & 972.6 & 1082.6 \\
        TP+Split Term & 733.9 & 1152.5 & 1232 \\
        Too Long & 143.5 & 201.3 & 233.7 \\
        Cut Off & 509.6 & 438.2 & 274.1 \\
        Precision & 0.420 & \textbf{0.652} & \textbf{0.697} \\
        Recall & 0.473 & 0.743 & 0.794 \\
        F$_1$ & 0.442 & 0.692 & \textbf{0.742} \\ \bottomrule
    \end{tabular}
    \caption{\label{tab:eval_10_fold_10240}Averaged performance of
    fine-tuned models, with 10\,240 training data samples}
\end{table}

\begin{table}[t]
    \centering
    \begin{tabular}{rrrr}
    \toprule
        \textbf{Model} & \textbf{cc\_ep01} & \textbf{cc\_ep10} & \textbf{Rob.} \\ \midrule
        2048 & 0.044 & 0.052 & \textbf{0.031} \\
        10240 & 0.043 & 0.026 & \textbf{0.011} \\ \bottomrule
    \end{tabular}
    \caption{\label{tab:eval_10_fold_std_f1}The standard deviation of the
    F$_1$ score of different fine-tuned models, with 2048 and with 10\,240
  training data samples}
\end{table}

\section{Conclusion}
In this work, we have contributed to the efficient creation of a labeled
dataset for definiendum extraction from mathematical papers. We have then
compared two usages of transformers:  asking GPT vs fine-tuning
pre-trained language models. Our experimental results show GPT-4's
capacity to understand mathematical texts with only one example in the
prompt. We highlight the good precision--recall balance and the relatively low cost of fine-tuning Roberta for this domain-specific information extraction task. A constraint of our work comes from the nature of our labeled data because authors have their own writing styles: there could be more than one correct annotation for a phrase. For instance, our definition blocks are compiled from \LaTeX{} sources, and we plan to test our fine-tuned models on definitions extracted from real PDF format papers without \LaTeX{} sources. \citet{pluvinage_2020} proposes sentence-level classification and text segmentation to retrieve mathematical results from PDF and can provide a preliminary test set for us. For future work, we will explore the ambiguities of extracted entities and link them to classes. Our experience with cc\_math\_roberta models also open up research about improving the robustness over different NLP tasks of from-scratched domain-specific language models.

\section*{Acknowledgments}

This work was funded in part by the French government under
management of Agence Nationale de la Recherche as part of the
“Investissements d’avenir” program, reference ANR-19-P3IA-0001
(PRAIRIE 3IA Institute).

\bibliography{anthology,custom}

\end{document}